\def\year{2018}\relax
\documentclass[letterpaper]{article} 
\usepackage{aaai18}  
\usepackage{times}  
\usepackage{helvet}  
\usepackage{courier}  
\usepackage{url}  
\usepackage{graphicx}  

\usepackage{amsmath}
\usepackage{amssymb}
\usepackage{color}
\usepackage[]{algorithm2e}
\usepackage{multirow}
\usepackage{comment}

\begin{document}
%
\title{Interpreting CNN Knowledge Via An Explanatory Graph}
\author{Quanshi Zhang, Ruiming Cao, Feng Shi, Ying Nian Wu, and Song-Chun Zhu\\University of California, Los Angeles}

\maketitle
\begin{abstract}
This paper learns a graphical model, namely an explanatory graph, which reveals the knowledge hierarchy hidden inside a pre-trained CNN. Considering that each filter\footnote[1]{The output of a conv-layer is called the feature map of a conv-layer. Each channel of this feature map is produced by a filter, so we call a channel the feature map of a filter.} in a conv-layer of a pre-trained CNN usually represents a mixture of object parts, we propose a simple yet efficient method to automatically disentangles different part patterns from each filter, and construct an explanatory graph. In the explanatory graph, each node represents a part pattern, and each edge encodes co-activation relationships and spatial relationships between patterns. More importantly, we learn the explanatory graph for a pre-trained CNN in an unsupervised manner, \emph{i.e.} without a need of annotating object parts. Experiments show that each graph node consistently represents the same object part through different images. We transfer part patterns in the explanatory graph to the task of part localization, and our method significantly outperforms other approaches.
\end{abstract}

\section{Introduction}

Convolutional neural networks (CNNs)~\cite{CNN,CNNImageNet,ResNet,ConvFace} have achieved superior performance in object classification and detection. However, the end-to-end learning strategy makes the entire CNN a black box. When a CNN is trained for object classification, we believe that its conv-layers have encoded rich implicit patterns (\emph{e.g.} patterns of object parts and patterns of textures). Therefore, in this research, we aim to provide a global view of how visual knowledge is organized in a pre-trained CNN, which presents considerable challenges. For example,
\begin{itemize}
\item[1] How many types of patterns are memorized by each convolutional filter of the CNN (here, a pattern may describe a specific object part or a certain texture)?
\item[2] Which patterns are co-activated to describe an object part?
\item[3] What is the spatial relationship between two patterns?
\end{itemize}

\begin{figure}[t]
\centering
\includegraphics[width=0.99\linewidth]{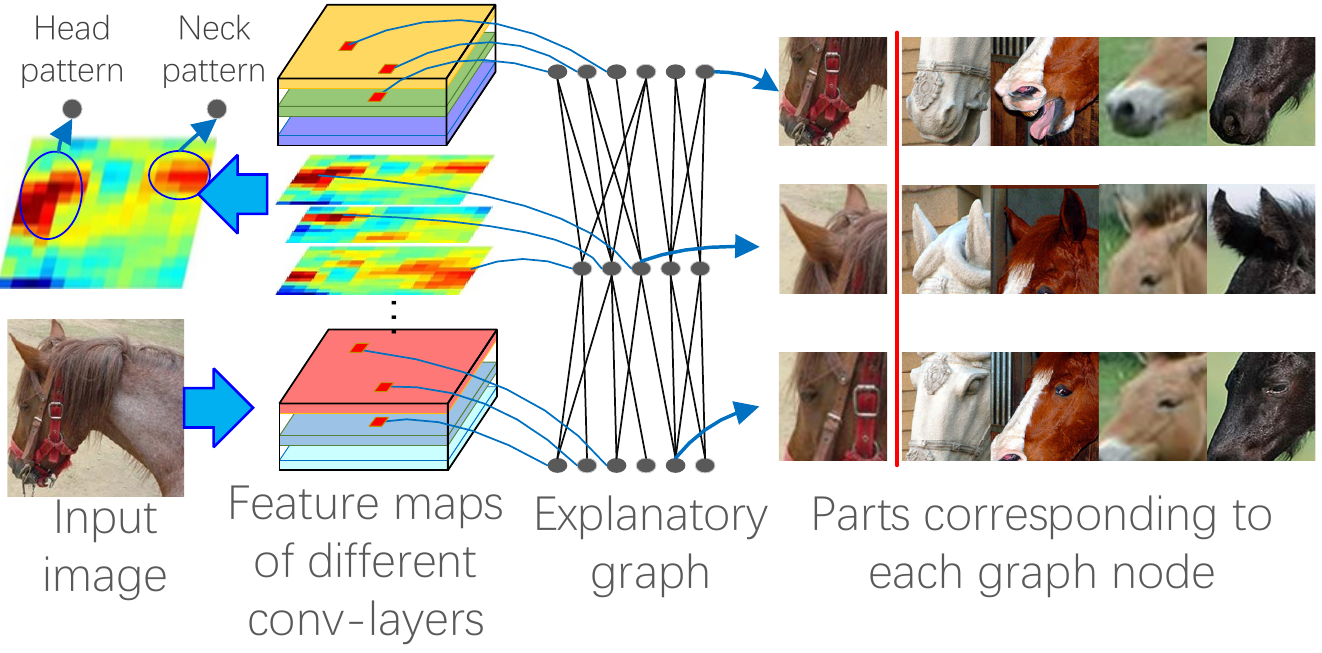}
\caption{An explanatory graph represents knowledge hierarchy hidden in conv-layers of a CNN. Each filter in a pre-trained CNN may be activated by different object parts. Our method disentangles part patterns from each filter in an unsupervised manner, thereby clarifying the knowledge representation.}
\label{fig:top}
\end{figure}

In this study, given a pre-trained CNN, we propose to mine mid-level object part patterns from conv-layers, and we organize these patterns in an explanatory graph in an unsupervised manner. As shown in Fig.~\ref{fig:top}, the explanatory graph explains the knowledge hierarchy hidden inside the CNN. The explanatory graph disentangles the mixture of part patterns in each filter's feature map\textcolor{red}{\footnotemark[1]} of a conv-layer, and uses each graph node to represent a part.

\noindent\textbf{\textbullet\quad Representing knowledge hierarchy:} The explanatory graph has multiple layers, which correspond to different conv-layers of the CNN. Each graph layer has many nodes. We use these graph nodes to summarize the knowledge hidden in chaotic feature maps of the corresponding conv-layer. Because each filter in the conv-layer may potentially represent multiple parts of the object, we use graph nodes to represent patterns of all candidate parts. A graph edge connects two nodes in adjacent layers to encode co-activation logics and spatial relationships between them.

Note that we do \textbf{not} fix the location of each pattern (node) to a certain neural unit of a conv-layer's output. Instead, given different input images, a part pattern may appear on various positions of a filter's feature maps\textcolor{red}{\footnotemark[1]}. For example, the horse face pattern and the horse ear pattern in Fig.~\ref{fig:top} can appear on different positions of different images, as long as they are co-activated and keep certain spatial relationships.

\noindent\textbf{\textbullet\quad Disentangling object parts from a single filter:} As shown in Fig.~\ref{fig:top}, each filter in a conv-layer may be activated by different object parts (\emph{e.g.} the filter's feature map\textcolor{red}{\footnotemark[1]} may be activated by both the head and the neck of a horse). To clarify the knowledge representation, we hope to disentangle patterns of different object parts from the same filter in an unsupervised manner, which presents a big challenge for state-of-the-art algorithms.

In this study, we propose a simple yet effective method to automatically discover object parts from a filter's feature maps \textbf{without} ground-truth part annotations. In this way, we can filter out noisy activations from feature maps, and we ensure that each graph node consistently represents the same object part among different images.

Given a testing image to the CNN, the explanatory graph can tell 1) whether a node (part) is triggered and 2) the location of the part on the feature map.

\noindent\textbf{\textbullet\quad Graph nodes with high transferability:} Just like a dictionary, the explanatory graph provides off-the-shelf patterns for object parts, which enables a probability of transferring knowledge from conv-layers to other tasks. Considering that all filters in the CNN are learned using numerous images, we can regard each graph node as a detector that has been sophisticatedly learned to detect a part among thousands of images. Compared to chaotic feature maps of conv-layers, our explanatory graph is a more concise and meaningful representation of the CNN knowledge.

To prove the above assertions, we learn explanatory graphs for different CNNs (including the VGG-16, residual networks, and the encoder of a VAE-GAN) and analyze the graphs from different perspectives as follows.

\noindent\textit{\bf Visualization \& reconstruction:} Patterns in graph nodes can be directly visualized in two ways. First, for each graph node, we list object parts that trigger strong node activations. Second, we use activation states of graph nodes to reconstruct image regions related to the nodes.

\noindent\textit{\bf Examining part interpretability of graph nodes:} \cite{Interpretability} defined different types of interpretability for a CNN. In this study, we evaluate the part-level interpretability of the graph nodes. \emph{I.e.} given an explanatory graph, we check whether a node consistently represents the same part semantics among different objects. We follow ideas of \cite{Interpretability,CNNSemanticDeep} to measure the part interpretability of each node.

\noindent\textit{\bf Examining location instability of graph nodes:} Besides the part interpretability, we also define a new metric, namely location instability, to evaluate the clarity of the semantic meaning of each node in the explanatory graph. We assume that if a graph node consistently represents the same object part, then the distance between the inferred part and some ground-truth semantic parts of the object should not change a lot among different images.

\noindent\textit{\bf Testing transferability:} We associate graph nodes with explicit part names for multi-shot part localization. The superior performance of our method shows the good transferability of our graph nodes.

In experiments, we demonstrate both the representation clarity and the high transferability of the explanatory graph.

\noindent\textbf{Contributions} of this paper are summarized as follows.

\noindent
1) In this paper, we, for the first time, propose a simple yet effective method to clarify the chaotic knowledge hidden inside a pre-trained CNN and to summarize such a deep knowledge hierarchy using an explanatory graph. The graph disentangles part patterns from each filter of the CNN. Experiments show that each graph node consistently represents the same object part among different images.

\noindent
2) Our method can be applied to different CNNs, \emph{e.g.} VGGs, residual networks, and the encoder of a VAE-GAN.

\noindent
3) The mined patterns have good transferability, especially in multi-shot part localization. Although our patterns were pre-trained without part annotations, our transfer-learning-based part localization outperformed approaches that learned part representations with part annotations.

\section{Related work}

\subsection{Semantics in CNNs}

The interpretability and the discrimination power are two crucial aspects of a CNN~\cite{Interpretability}. In recent years, different methods are developed to explore the semantics hidden inside a CNN. Many statistical methods~\cite{CNNAnalysis_1,CNNAnalysis_2,CNNVisualization_5} have been proposed to analyze the characteristics of CNN features. In particular, \cite{CNNBias} has demonstrated that in spite of the good classification performance, a CNN may encode biased knowledge representations due to dataset bias. Instead, the CNN usually uses unreliable contexts for classification. For example, a CNN may extract features from hairs as a context to identify the \textit{smiling} attribute. Therefore, we need methods to visualize the knowledge hierarchy hidden inside a CNN.

\textbf{Visualization \& interpretability of CNN filters:}{\verb| |} Visualization of filters in a CNN is the most direct way of exploring the pattern hidden inside a neural unit. Up-convolutional nets~\cite{FeaVisual} were developed to invert feature maps to images. Comparatively, gradient-based visualization~\cite{CNNVisualization_1,CNNVisualization_2,CNNVisualization_3} showed the appearance that maximized the score of a given unit, which is more close to the spirit of understanding CNN knowledge. Furthermore, Bau \emph{et al.}~\cite{Interpretability} defined and analyzed the interpretability of each filter.

Although these studies achieved clear visualization results, theoretically, gradient-based visualization methods visualize one of the local minimums contained in a high-layer filter. \emph{I.e.} when a filter represents multiple patterns, these methods selectively illustrated one of the patterns; otherwise, the visualization result will be chaotic. Similarly, \cite{Interpretability} selectively analyzed the semantics among the highest 0.5\% activations of each filter. In contrast, our method provides a solution to explaining both strong and weak activations of each filter and discovering all possible patterns from each filter.

\textbf{Pattern retrieval:}{\verb| |} Some studies go beyond passive visualization and actively retrieve units from CNNs for different applications. Like middle-level feature extraction~\cite{MiddleLevel}, pattern retrieval mainly learns mid-level representations of CNN knowledge. Zhou~\emph{et al.}~\cite{CNNSemanticDeep,CNNSemanticDeep2} selected units from feature maps to describe ``scenes''. Simon~\emph{et al.} discovered objects from feature maps of unlabeled images~\cite{ObjectDiscoveryCNN_2}, and selected a filter to describe each part in a supervised fashion~\cite{CNNSemanticPart}. However, most methods simply assumed that each filter mainly encoded a single visual concept, and ignored the case that a filter in high conv-layers encoded a mixture of patterns. \cite{CNNAoG,DeepQA,interactiveAOG_arXiv} extracted certain neurons from a filter's feature map to describe an object part in a weakly-supervised manner (\emph{e.g.} learning from active question answering and human interactions).

In this study, the explanatory graph disentangles patterns different parts in the CNN without a need of part annotations. Compared to raw feature maps, patterns in graph nodes are more interpretable.

\subsection{Weakly-supervised knowledge transferring}

Knowledge transferring ideas have been widely used in deep learning. Typical research includes end-to-end fine-tuning and transferring CNN knowledge between different categories~\cite{CNNAnalysis_2} or different datasets~\cite{UnsuperTransferCNN}. In contrast, we believe that a transparent representation of part knowledge will create a new possibility of transferring part knowledge to other applications. Therefore, we build an explanatory graph to represent part patterns hidden inside a CNN, which enables transfer part patterns to other tasks. Experiments have demonstrated the efficiency of our method in multi-shot part localization.

\section{Algorithm}

\subsection{Intuitive understanding of the pattern hierarchy}
\label{sec:pre}

\begin{figure}[t]
\centering
\includegraphics[width=0.99\linewidth]{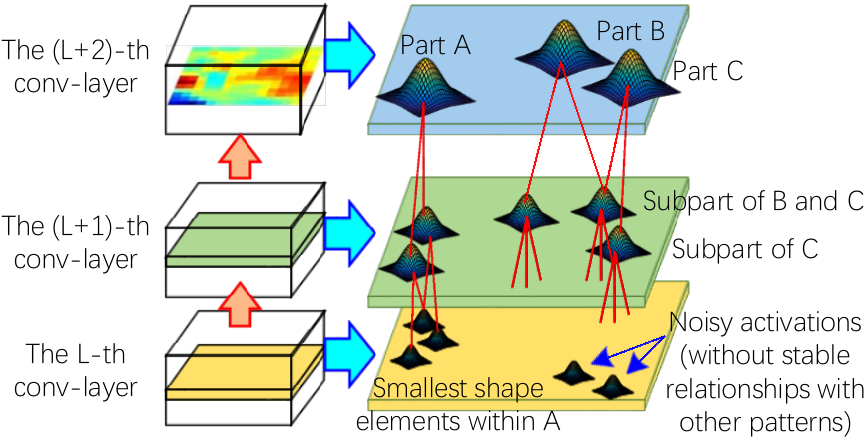}
\caption{Spatial and co-activation relationships between part patterns in the explanatory graph. High-layer patterns filter out noises and disentangle low-layer patterns. From another perspective, we can regard low-layer patterns as components of high-layer patterns.}
\label{fig:peak}
\end{figure}

As shown in Fig.~\ref{fig:peak}, the feature map of a filter can usually be activated by different object parts in various locations. Let us assume that a feature map is activated with $N$ peaks. Some peaks represent common parts of the object, and we call such activation peaks \textit{part patterns}. Whereas, other peaks may correspond to background noises.

Our task is to discover activation peaks of part patterns out of noisy peaks from a filter's feature map. We assume that if a peak corresponds to an object part, then some patterns of other filters must be activated in similar map positions; vice versa. These patterns represent sub-regions of the same part and keep certain spatial relationships. Thus, in the explanatory graph, we connect each pattern in a low conv-layer to some patterns in the neighboring upper conv-layer. We mine part patterns layer by layer. Given patterns mined from the upper conv-layer, we select activation peaks, which keep stable spatial relationships with specific upper-layer patterns among different images, as part patterns in the current conv-layer.

As shown in Fig.~\ref{fig:peak}, patterns in high conv-layers usually represent large-scale object parts. Whereas, patterns in low conv-layers mainly describes relatively simple shapes, which are less distinguishable in semantics. We use high-layer patterns to filter out noises and disentangle low-layer patterns. From another perspective, we can regard low-layer patterns as components of high-layer patterns.

\subsection{Learning}

\textbf{Notations:} We are given a CNN pre-trained using its own set of training samples {\small${\bf I}$}. Let $G$ denote the target explanatory graph. $G$ contains several layers, which corresponds to conv-layers in the CNN. We disentangles the $d$-th filter of the $L$-th conv-layer into {\small$N_{L,d}$} different part patterns, which are modeled as a set of {\small$N_{L,d}$} nodes in the $L$-th layer of $G$, denoted by {\small$\Omega_{L}$}. {\small$\Omega_{L,d}\subset\Omega_{L}$} denotes the node set for the $d$-th filter. Parameters of these nodes in the $L$-th layer are given as {\small${\boldsymbol\theta}_{L}$}, which mainly encode spatial relationships between these nodes and the nodes in the $(L+1)$-th layer.

Given a training image {\small$I\in{\bf I}$}, the CNN generates a feature map\textcolor{red}{\footnotemark[1]} of the $L$-th conv-layer, denoted by {\small${\bf X}_{L}^{I}$}. Then, for each node {\small$V\in\Omega_{L,d}$}, we can use the explanatory graph to infer whether $V$'s part pattern appears on the $d$-th channel\footnotemark[1] of {\small${\bf X}_{L}^{I}$}, as well as the position of the part pattern (if the pattern appears). We use {\small${\bf R}_{L}^{I}$} to represent position inference results for all nodes in the $L$-th layer.

\textbf{Objective function:} We build the explanatory graph in a top-down manner. Given all training samples {\small${\bf I}$}, we first disentangle patterns from the top conv-layer of the CNN, and built the top graph layer. Then, we use inference results of the patterns/nodes on the top layer to help disentangle patterns from the neighboring lower conv-layer. In this way, the construction of $G$ is implemented layer by layer. Given inference results for the $(L+1)$-th layer {\small$\{{\bf R}_{L+1}^{I}\}_{I\in{\bf I}}$}, we expect that all patterns to simultaneously 1) be well fit to {\small${\bf X}_{L}^{I}$} and 2) keep consistent spatial relationships with upper-layer patterns {\small${\bf R}_{L+1}^{I}$} among different images. The objective of learning for the $L$-th layer is given as
\begin{equation}
{\arg\!\max}_{{\boldsymbol\theta}_{L}}{\prod}_{I\in{\bf I}}P({\bf X}_{L}^{I}|{\bf R}_{L+1}^{I},{\boldsymbol\theta}_{L})
\label{eqn:prob}
\end{equation}
\emph{I.e.} we learn node parameters {\small${\boldsymbol\theta}_{L}$} that best fit feature maps of training images.

Let us focus on a conv-layer's feature map {\small${\bf X}_{L}^{I}$} of image $I$. Without ambiguity, we ignore the superscript $I$ to simplify notations in following paragraphs. We can regard {\small${\bf X}_{L}$} as a distribution of ``neural activation entities.'' We consider the neural response of each unit {\small$x\in{\bf X}_{L}$} as the number of ``activation entities.'' In other words, each unit $x$ localizes at the position of {\small${\bf p}_{x}$}\footnote[2]{To make unit positions in different conv-layers comparable with each other (\emph{e.g.} {\small$\mu_{V'\rightarrow V}$} in Eq.~\ref{eqn:gauss}), we project the position of unit $x$ to the image plane. We define the coordinate {\small${\bf p}_{x}$} on the image plane, instead of on the feature-map plane.} in the $d_{x}$-th channel of {\small${\bf X}_{L}$}. We use {\small$F(x)\!=\!\beta\cdot\max\{f_{x},0\}$} to denote the number of activation entities at the position {\small${\bf p}_{x}$}, where $f_{x}$ is the normalized response value of $x$; $\beta$ is a constant.

\begin{figure}[t]
\centering
\includegraphics[width=0.9\linewidth]{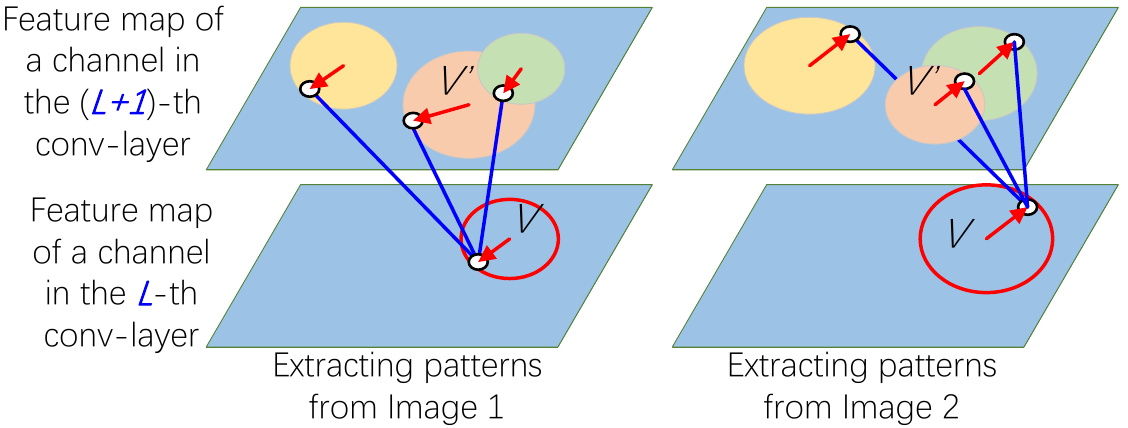}
\caption{Related patterns $V$ and $V'$ keep similar spatial relationships among different images. Circle centers represent the prior pattern positions, \emph{e.g.} $\mu_{V}$ and $\mu_{V'}$. Red arrows denote relative displacements between the inferred positions and prior positions, \emph{e.g.} ${\bf p}_{V}-\mu_{V}$.}
\label{fig:pair}
\end{figure}

Therefore, just like a Gaussian mixture model, we use all patterns in {\small$\Omega_{L,d}$} as a mixture model to jointly explain the distribution of activation entities on the $d$-th channel of {\small${\bf X}_{L}$}:
\begin{small}
\begin{eqnarray}
\!\!\!\!&P({\bf X}_{L}|{\bf R}_{L+1},{\boldsymbol\theta}_{L})\!=\!{\prod}_{x\in{\bf X}_{L}}P({\bf p}_{x}|{\bf R}_{L+1},{\boldsymbol\theta}_{L})^{F(x)}\\
\!\!\!\!&={\prod}_{x\in{\bf X}_{L}}\Big\{\sum\limits_{V\in\Omega_{L,d}\cup\{V_{\textrm{none}}\}}\!\!\!\!\!\!\!\!\!P(V)P({\bf p}_{x}|V,{\bf R}_{L+1},{\boldsymbol\theta}_{L})\!\Big\}_{d=d_{x}}^{F(x)}\!\nonumber
\end{eqnarray}
\end{small}
where we consider each node {\small$V\in\Omega_{L,d}$} as a hidden variable or an alternative component in the mixture model to describe activation entities. {\small$P(V)=\frac{1}{N_{L,d}+1}$} is a constant prior probability. {\small$P({\bf p}_{x}|V,{\bf R}_{L+1},{\boldsymbol\theta}_{L})$} measures the compatibility of using node $V$ to describe an activation entity at ${\bf p}_{x}$. In addition, because noisy activations cannot be explained by any patterns, we add a dummy component {\small$V_{\textrm{none}}$} to the mixture model for noisy activations. Thus, the compatibility between $V$ and ${\bf p}_{x}$ is computed based on spatial relationship between $V$ and other nodes in $G$, which is roughly formulated as
\begin{small}
\begin{eqnarray}
\!\!\!\!P({\bf p}_{x}|V,{\bf R}_{L+1},{\boldsymbol\theta}_{L})\!=\!\left\{\!\begin{array}{ll}\gamma\!\!\!\!\!\prod\limits_{V'\in{E}_{V}}\!\!\!P({\bf p}_{x}|{\bf p}_{V'},{\boldsymbol\theta}_{L})^{\lambda}\!\!,\!\!\!&\!\!\!V\!\in\!\Omega_{L\!,d_{x}}\!\!\!\!\!\!\!\!\!\!\!\!\!\!\\
\gamma\tau,&\!\!\!\!\!\!\!\!\!\!\!\!V\!=\!V_{\textrm{none}}\!\!\!\!\!\!\!\!\!\!\!\!\\
\end{array}\right.\label{eqn:prob-full}\\
\!\!\!P({\bf p}_{x}|{\bf p}_{V'},{\boldsymbol\theta}_{L})\!=\!{\bf\mathcal N}({\bf p}_{x}|\mu_{V'\!\rightarrow\!V},\sigma_{V'}^2)\!\!\label{eqn:gauss}
\end{eqnarray}
\end{small}
In above equations, node $V$ has a set of $M$ neighboring patterns in the upper layer, denoted by {\small${E}_{V}\in{\boldsymbol\theta}_{L}$}, which would be determined during the learning process. The overall compatibility {\small$P({\bf p}_{x}|V,{\bf R}_{L+1},{\boldsymbol\theta}_{L})$} is divided into the spatial compatibility between node $V$ and each neighboring node $V'$, {\small$P({\bf p}_{x}|{\bf p}_{V'},{\boldsymbol\theta}_{L})$}. {\small$\forall V'\in{E}_{V}$}, {\small${\bf p}_{V'}\!\in\!{\bf R}_{L+1}$} denotes the position inference result of $V'$, which have been provided. {\small$\lambda=\frac{1}{M}$} is a constant for normalization. {\small$\gamma$} is a constant to roughly ensure {\small$\int P({\bf p}_{x}|V,{\bf R}_{L+1},{\boldsymbol\theta}_{L}){\bf d}{{\bf p}_{x}}=1$}, which can be eliminated during the learning process.

\begin{algorithm}[t]
{\bf Inputs:} feature map ${\bf X}_{L}$ of the $L$-th conv-layer, inference results ${\bf R}_{L+1}$ in the upper conv-layer.\\
{\bf Outputs:} $\mu_{V},{E}_{V}$ for $\forall V\in\Omega_{L}$.\\
{\bf Initialization:} $\forall V$, ${E}_{V}\!=\!\{V_{\textrm{dummy}}\}$, a random value for $\mu_{V}^{(0)}$\\
\For{$iter=1$ to $T$}{
$\forall V\in\Omega_{L}$, compute $P({\bf p}_{x},V|{\bf R}_{L+1},{\boldsymbol\theta}_{L})$.\\
\For{$V\in\Omega_{L}$}{
1) Update $\mu_{V}$ via an EM algorithm,\\
{\small$\mu_{V}^{(iter)}\!=\!\mu_{V}^{(iter-1)}\!\!+\!\eta\!\!\!\!\!\!\!\sum\limits_{I\in{\bf I},x\in{\bf X}_{L}}\!\!\!\!\!\!\!{\bf\large E}_{P(V|{\bf p}_{x},{\bf R}_{L+1},{\boldsymbol\theta}_{L})}\big[$ $F(x)\cdot\frac{\partial{\log}P({\bf p}_{x},V|{\bf R}_{L+1},{\boldsymbol\theta}_{L})}{\partial\mu_{V}}\big]$}.\\
2) Select $M$ patterns from $V'\in\Omega_{L+1}$ to construct ${E}_{V}$ based on a greedy strategy, which maximize {\small${\prod}_{I\in{\bf I}}P({\bf X}_{L}|{\bf R}_{L+1},{\boldsymbol\theta}_{L})$}.}}
\caption{Learning sub-graph in the $L$-th layer}
\label{alg:main}
\end{algorithm}

As shown in Fig.~\ref{fig:pair}, an intuitive idea is that the relative displacement between $V$ and $V'$ should not change a lot among different images. Let {\small$\mu_{V}\in{\boldsymbol\theta}_{L}$} and {\small$\mu_{V'}\in{\boldsymbol\theta}_{L+1}$} denote the prior positions of $V$ and $V'$, respectively. Then, {\small${\bf p}_{x}-{\bf p}_{V'}$} will approximate to {\small$\mu_{V}-\mu_{V'}$}, if node $V$ can well fit activation entities at ${\bf p}_{x}$. Therefore, given {\small${E}_{V}$} and {\small${\bf R}_{L+1}$}, we assume the spatial relationship between $V$ and $V'$ follows a Gaussian distribution in Eqn.~\ref{eqn:gauss}, where {\small$\mu_{V'\rightarrow V}\!=\!\mu_{V}-\mu_{V'}+{\bf p}_{V'}$} denotes the prior position of $V$ given $V'$. {\small$\sigma_{V'}^2$} denotes the variation, which can be estimated from data\footnote[3]{We can prove that for each $V\in\Omega_{L,d}$, $P({\bf p}_{x}|V,{\bf R}_{L+1},{\boldsymbol\theta}_{L})$ $\propto{\bf\mathcal N}({\bf p}_{x}|\mu_{V}$ $+\Delta_{I,V},\tilde{\sigma}_{V}^2)$, where $\Delta_{I,V}=\sum_{V'\in{E}_{V}}$ $\frac{{\bf p}_{V'}-\mu_{V'}}{\sigma_{V'}^2}$ $/\sum_{V'\in{E}_{V}}\frac{1}{\sigma_{V'}^2}$; $\tilde{\sigma}_{V}^2$ $=1/{\bf E}_{V'\in{E}_{V}}\frac{1}{\sigma_{V'}^2}$. Therefore, we can either directly use $\tilde{\sigma}_{V}^2$ as $\sigma_{V}^2$, or compute the variation of ${\bf p}_{x}-\mu_{V}-\Delta_{I,V}$ \emph{w.r.t.} different images to obtain $\sigma_{V}^2$.}.

In this way, the core of learning is to determine an optimal set of neighboring patterns {\small${E}_{V}\in{\boldsymbol\theta}_{L}$} and estimate the prior position {\small$\mu_{V}\in{\boldsymbol\theta}_{L}$}. Note that our method only models the relative displacement {\small$\mu_{V}-\mu_{V'}$}.

\begin{figure}[t]
\centering
\includegraphics[width=0.8\linewidth]{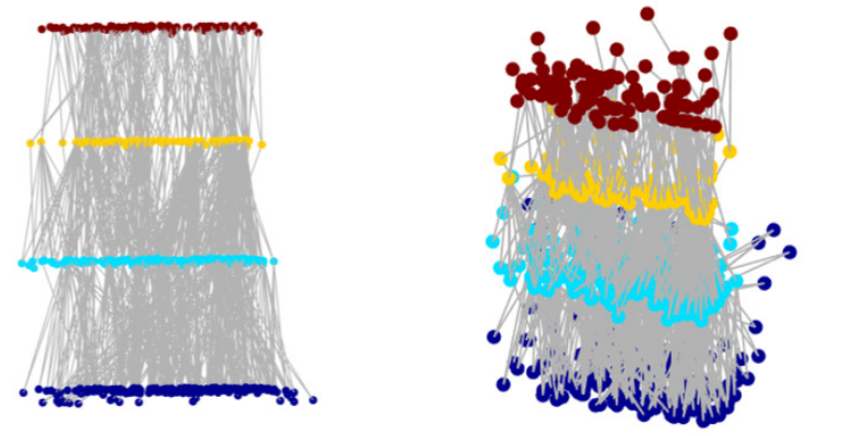}
\caption{A four-layer explanatory graph. For clarity, we put all nodes of different filters in the same conv-layer on the same plan and only show 1\% of the nodes with 10\% of their edges from two perspectives.}
\label{fig:global}
\end{figure}

\begin{figure*}[t]
\centering
\includegraphics[width=0.86\linewidth]{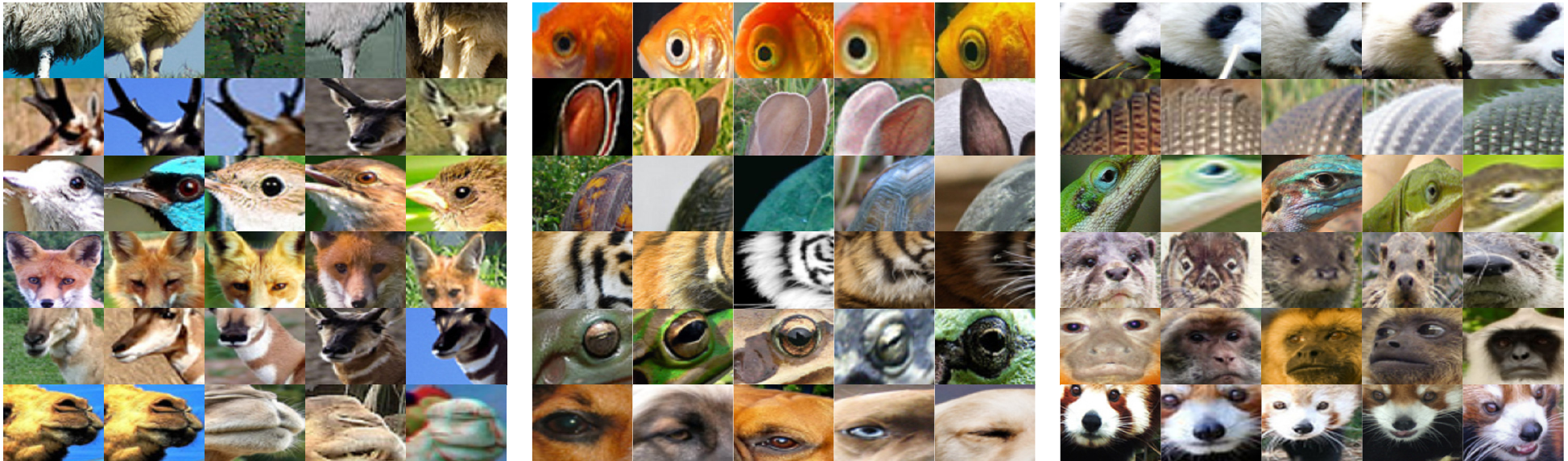}
\caption{Image patches corresponding to different nodes in the explanatory graph.}
\label{fig:patch}
\end{figure*}

\textbf{Inference of pattern positions:} Given the $d$-th filter's feature map, we simply assign node {\small$V\in\Omega_{L,d}$} with a certain unit {\small$\hat{x}={\arg\!\max}_{x\in{\bf X}_{L}:d_{x}=d}S_{V\rightarrow x}^{I}$} on the feature map as the true inference of $V$, where {\small$S_{V\rightarrow x}^{I}\!=\!F(x)P({\bf p}_{x}|V,{\bf R}_{L+1},{\boldsymbol\theta}_{L})$} denotes the score of assigning $V$ to $x$. Accordingly, {\small${\bf p}_{V'}={\bf p}_{\hat{x}}$} represents the inferred position of $V$. In particular, in Eqn.~(\ref{eqn:prob}), we define {\small${\bf R}_{L+1}\!=\!\{{\bf p}_{V'}\}_{V'\!\in\!\Omega_{L+1}}$}.

\textbf{Top-down EM-based Learning:}{\verb| |} For each node $V$, we need to learn the parameter {\small$\mu_{V}\in{\boldsymbol\theta}_{L}$} and a set of patterns in the upper layer that are related to $V$, {\small${E}_{V}\!\in\!{\boldsymbol\theta}_{L}$}. We learn the model in a top-down manner. We first learn nodes in the top-layer of $G$, and then learn for the neighboring lower layer. For the sub-graph in the $L$-th layer, we iteratively estimate parameters of {\small$\mu_{V}$} and {\small${E}_{V}$} for nodes in the sub-graph. We can use the Expectation-Maximization (EM) algorithm for learning. Please see Algorithm~\ref{alg:main} for details.

Note that for each pattern $V$ in the top conv-layer, we simply define {\small${E}_{V}\!=\!\{V_{\textrm{dummy}}\}$}, In {\small${\bf R}_{L+1}$}, {\small$\mu_{V_{\textrm{dummy}}}\!=\!{\bf p}_{V_{\textrm{dummy}}}\!=\!{\bf 0}$}. {\small$V_{\textrm{dummy}}$} is a dummy node. Based on Eqns.~(\ref{eqn:prob-full}) and (\ref{eqn:gauss}), we obtain {\small$P({\bf p}_{x}|V,{\bf R}_{L+1},{\boldsymbol\theta}_{L})\!=\!{\bf\mathcal N}({\bf p}_{x}|\mu_{V},\sigma_{V}^2)$}.

\section{Experiments}

\subsection{Overview of experiments}

\textbf{Four types of CNNs:} To demonstrate the broad applicability of our method, we applied our method to four types of CNNs, \emph{i.e.} the VGG-16~\cite{VGG}, the 50-layer and 152-layer Residual Networks~\cite{ResNet}, and the encoder of the VAE-GAN~\cite{VAEGAN}.

\textbf{Three experiments and thirteen baselines:} We designed three experiments to evaluate the explanatory graph. The first experiment is to visualize patterns in the graph. The second experiment is to evaluate the semantic interpretability of the part patterns, \emph{i.e.} checking whether a pattern consistently represents the same object region among different images. We compared our patterns with three types of middle-level features and neural patterns. The third experiment is multi-shot learning for part localization, in order to test the transferability of patterns in the graph. In this experiment, we associated part patterns with explicit part names for part localization. We compared our method with ten baselines.

\textbf{Three benchmark datasets:} We built explanatory graphs to describe CNNs learned using a total of 37 animal categories in three datasets: the ILSVRC 2013 DET Animal-Part dataset~\cite{CNNAoG}, the CUB200-2011 dataset~\cite{CUB200}, and the Pascal VOC Part dataset~\cite{SemanticPart}. As discussed in \cite{SemanticPart,CNNAoG}, animals usually contain non-rigid parts, which presents a key challenge for part localization. Thus, we selected animal categories in the three datasets for testing.

\subsection{Implementation details}

\begin{figure*}[t]
\centering
\includegraphics[width=1.0\linewidth]{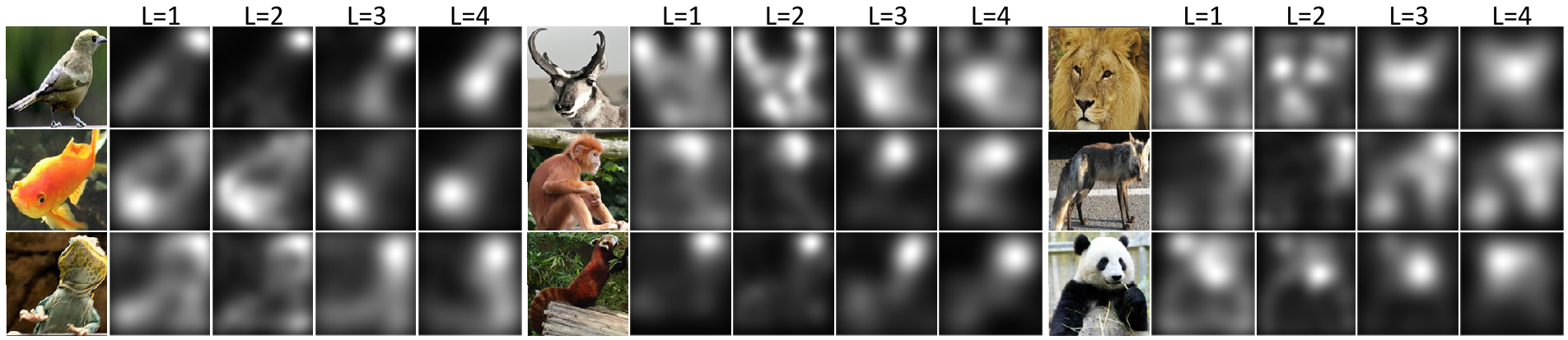}
\caption{Heat maps of patterns. We use a heat map to visualize the spatial distribution of the top-50\% patterns in the $L$-th layer of the explanatory graph with the highest inference scores.}
\label{fig:heatmap}
\end{figure*}

We first trained/fine-tuned a CNN using object images of a category, which were cropped using object bounding boxes. Then, we learned an explanatory graph to represent patterns of the category hidden inside the CNN. We set parameters {\small$\tau\!=\!0.1$}, {\small$M\!=\!15$}, {\small$T\!=\!20$}, and {\small$\beta\!=\!1$}.

\textbf{VGG-16:} Given a VGG-16 that was pre-trained using the 1.3M images in the ImageNet dataset~\cite{ImageNet}, we fine-tuned all conv-layers of the VGG-16 using object images in a category. The loss for finetuning was that for classification between the target category and background images. In each VGG-16, there are thirteen conv-layers and three fully connected layers. We selected the ninth, tenth, twelfth, and thirteenth conv-layers of the VGG-16 as four valid conv-layers, and accordingly built a four-layer graph. We extracted {\small$N_{L,d}$} patterns from the $d$-th filter of the $L$-th layer. We set {\small$N_{L=1\,\textrm{or}\,2,d}\!=\!40$} and {\small$N_{L=3\,\textrm{or}\,4,d}\!=\!20$}.

\textbf{Residual Networks:} We chose two residual networks, \emph{i.e.} the 50-layer and 152-layer ones. The finetuning process for each network was exactly the same as that for VGG-16. We built a three-layer graph based on each residual network by selecting the last conv-layer with a {\small$28\times28\times128$} feature ouput, the last conv-layer with a {\small$14\times14\times256$} feature map, and the last conv-layer with a {\small$7\times7\times512$} feature map as valid conv-layers. We set {\small$N_{L=1,d}\!=\!40$}, {\small$N_{L=2,d}\!=\!20$}, and {\small$N_{L=3,d}\!=\!10$}.

\textbf{VAE-GAN:} For each category, we used the cropped object images in the category to train a VAE-GAN. We learned a three-layer graph based on the three conv-layers of the encoder of the VAE-GAN. We set {\small$N_{L=1,d}\!=\!52$}, {\small$N_{L=2,d}\!=\!26$}, and {\small$N_{L=3,d}\!=\!13$}.

\subsection{Experiment 1: pattern visualization}

Given an explanatory graph for a VGG-16 network, we visualize its structure in Fig.~\ref{fig:global}. Part patterns in the graph are visualized in the following three ways.

\textbf{Top-ranked patches:} We performed pattern inference on all object images. For each image $I$, we extracted an images patch in the position of {\small${\bf p}_{\hat{x}_{V}}$}\footnote[4]{We projected the unit to the image to compute its position.} with a fixed scale of {\small$70\,pixels\!\times\!70\,pixels$} to represent pattern $V$. Fig.~\ref{fig:patch} shows a pattern's image patches that had highest inference scores.

\textbf{Heat maps of patterns:} Given a cropped object image $I$, we used the explanatory graph to infer its patterns on image $I$, and drew heat maps to show the spatial distribution of the inferred patterns. We drew a heat map for each layer $L$ of the graph. Given inference results of patterns in the $L$-th layer, we drew each pattern {\small$V\in\Omega_{L}$} as a weighted Gaussian distribution {\small$\alpha\cdot{\bf\mathcal N}(\mu\!=\!{\bf p}_{V},\sigma_{V}^2)$}\footnotemark[4] on the heat map, where {\small$\alpha\!=\!S_{V\rightarrow\hat{x}}^{I}$}. Please see Fig.~\ref{fig:heatmap} for heat maps of the top-50\% patterns with the highest scores of {\small$S_{V\rightarrow\hat{x}}^{I}$}.

\begin{figure}[t]
\centering
\includegraphics[width=\linewidth]{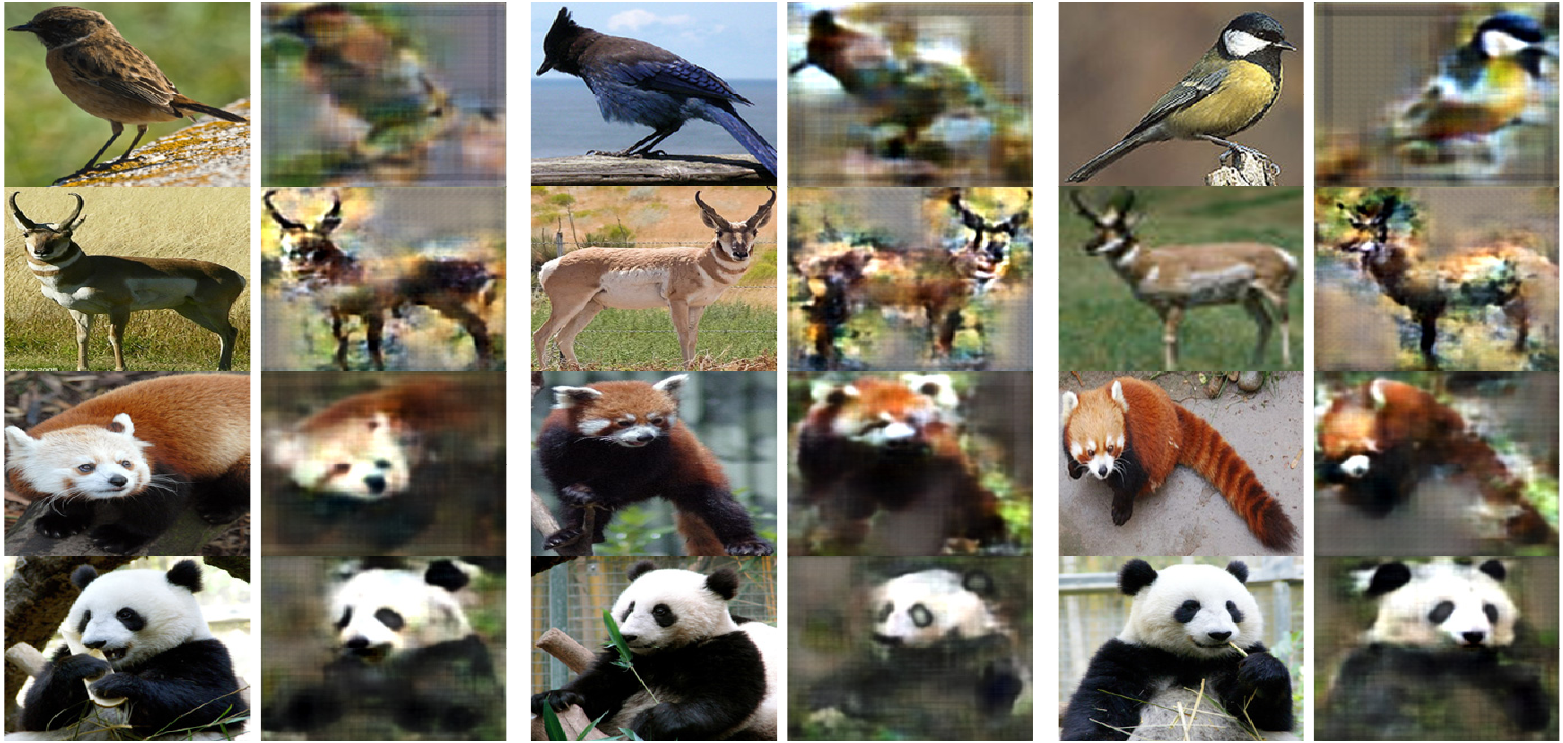}
\caption{Image synthesis result (right) based on patterns activated on an image (left). The explanatory graph only encodes major part patterns hidden in conv-layers, rather than compress a CNN without information loss. Synthesis results demonstrate that the patterns are automatically learned to represent foreground appearance, and ignore background noises and trivial details of objects.}
\label{fig:reconstruct}
\end{figure}

\textbf{Pattern-based image synthesis:} We used the up-convolutional network~\cite{FeaVisual} to visualize the learned patterns. Up-convolutional networks were originally trained for image reconstruction. In this study, given an image's feature maps corresponding to the second graph layer, we estimated the appearance of the original image. Given an object image $I$, we used the explanatory graph for pattern inference, \emph{i.e.} assigning each pattern $V$ with a certain neural unit {\small$\hat{x}_{V}$} as its position inference\textcolor{red}{\footnotemark[4]}. We considered the top-10\% patterns with highest scores of {\small$S_{V\rightarrow\hat{x}}^{I}$} as valid ones. We filtered out all neural responses of units, which were not assigned to valid patterns, from feature maps (setting these responses to zero). We then used \cite{FeaVisual} to synthesize the appearance corresponding to the modified feature maps. We regard image synthesis in Fig.~\ref{fig:reconstruct} as the visualization of the inferred patterns.

\begin{figure*}[t]
\centering
\includegraphics[width=0.95\linewidth]{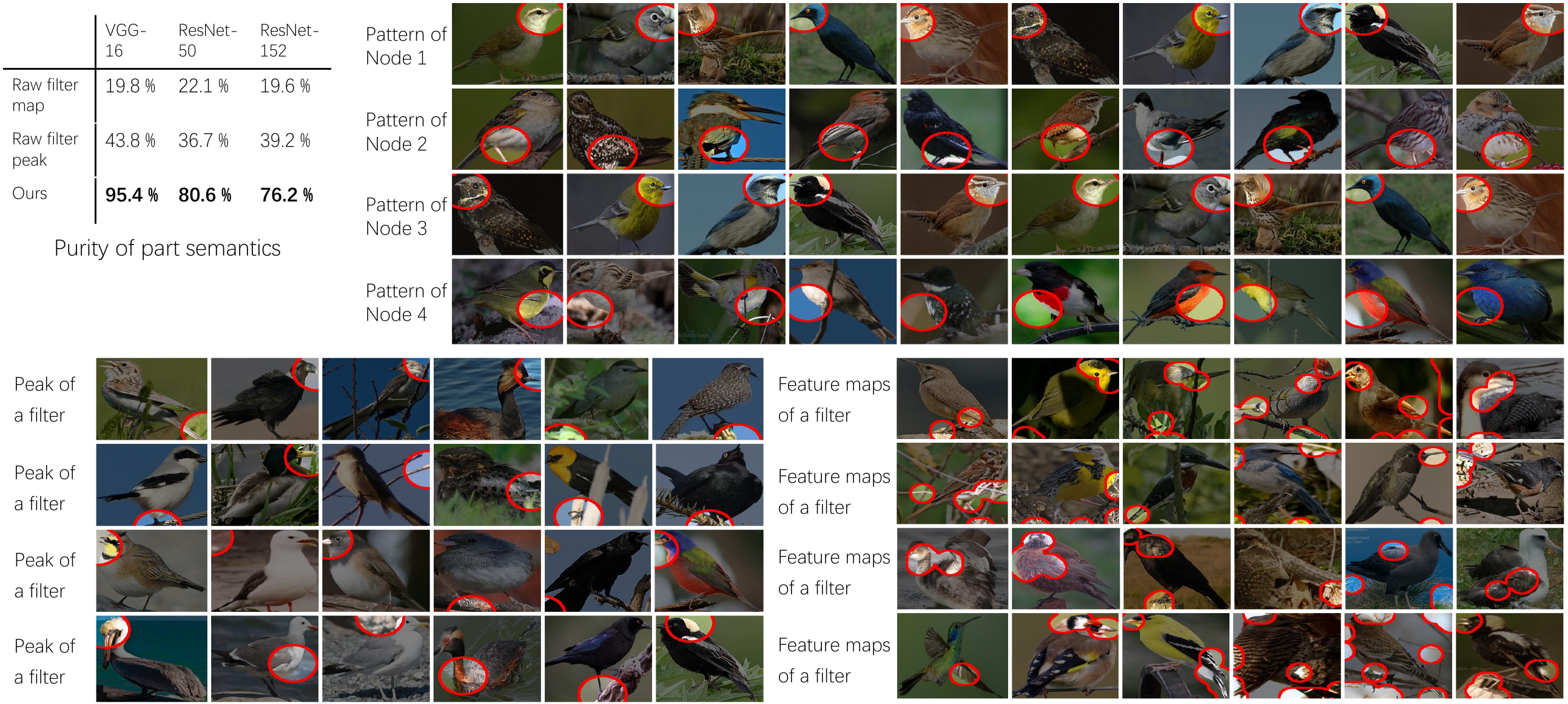}
\caption{Purity of part semantics. We draw image regions corresponding to each node in an explanatory graph and image regions corresponding to each pattern learned by other methods (we show some examples on the right). We use human users to annotate the semantic purity of each node/pattern. Cyan boxes show inference results that do not describe the common part.}
\label{fig:interpretability}
\end{figure*}

\subsection{Experiment 2: semantic interpretability of patterns}

In this experiment, we tested whether each pattern in an explanatory graph consistently represented the same object region among different images. We learned four explanatory graphs for a VGG-16 network, two residual networks, and a VAE-GAN that were trained/fine-tuned using the CUB200-2011 dataset~\cite{CUB200}. We used two methods to evaluate the semantic interpretability of patterns, as follows.

\textbf{Part interpretability of patterns:} We mainly extracted patterns from high conv-layers, and as discussed in \cite{Interpretability}, high conv-layers contain large-scale part patterns. We were inspired by Zhou \emph{et al.}~\cite{CNNSemanticDeep} and measured the interpretability of part patterns. For the pattern of a given node $V$, we used people to manually evaluate the pattern's interpretability. When we used $V$ to make inferences among all images, we regarded inference results with the top-$K$ inference scores $S_{V}^{I_{i}}$ among all images as valid representations of $V$. We require the $K$ highest inference scores $S_{V}^{I_{i}}$ on images $\{I_1,\ldots,I_{k}\}$ to take about 30\% of the inference energy, \emph{i.e.} {\small$\sum_{i=1}^{K}S_{V}^{I_{i}}=0.3\sum_{i\in{\bf I}}S_{V}^{I}$} (we use this equation to compute $K$). As shown in Fig.\ref{fig:interpretability}, we asked human raters how many inference results among the top $K$ described the same object part, in order to compute the purity of part semantics of pattern $V$.

The table in Fig.~\ref{fig:interpretability}(top-left) shows the semantic purity of the patterns in the second layer of the graph. Let the second graph layer correspond to the $L$-th conv-layer with $D$ filters. Like in \cite{CNNSemanticDeep}, the \textit{raw filter maps} baseline used activated neurons in the feature map of a filter to describe a part. The \textit{raw filter peaks} baseline considered the highest peak on a filer's feature map as a part detection. Like our method, the two baselines only visualized top-$K'$ part inferences (the $K'$ feature maps' neural activations took 30\% of activation energies among all images). We back-propagated the center of the receptive field of each neural activation to the image plane and simply used a fixed radius to draw the image region corresponding to each neural activation. Fig.~\ref{fig:interpretability} compares the image region corresponding to each node in the explanatory graph and image regions corresponding to feature maps of each filter. Our graph nodes encoded much more meaningful part representations than raw filters.

\begin{figure}[t]
\centering
\includegraphics[width=1.0\linewidth]{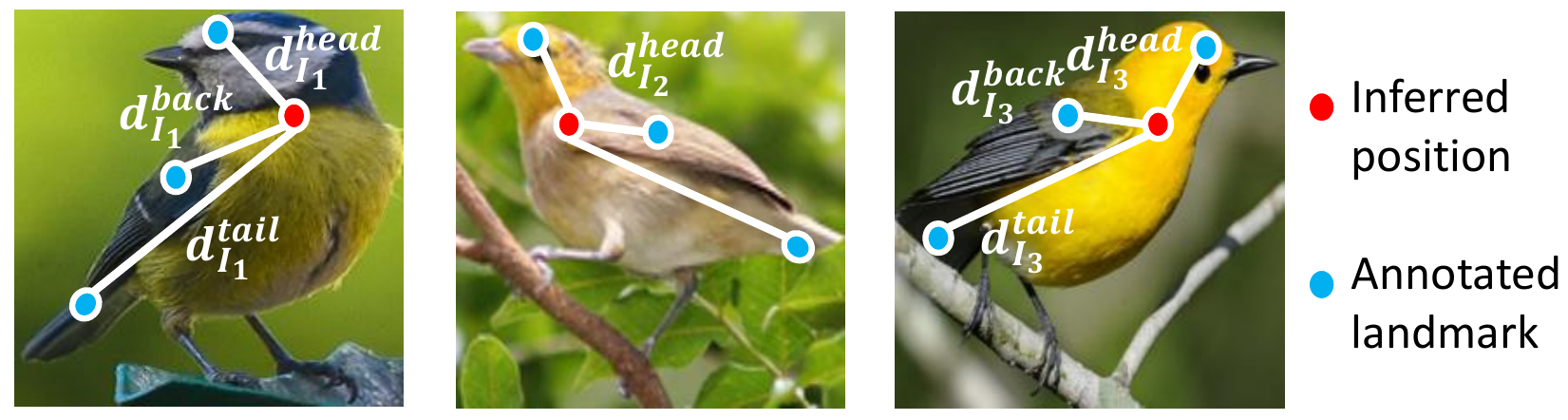}
\caption{Notation for the computation of location instability.}
\label{fig:instability}
\end{figure}

Because the baselines simply averaged the semantic purity among the $D$ filters, for a fair comparison, we also computed average semantic purities using the top-$D$ nodes, each node $V$ having the highest scores of $\sum_{i\in{\bf I}}S_{V}^{I}$.

\begin{table}[t]
\centering
\resizebox{1.0\linewidth}{!}{\begin{tabular}{l|cccc}
\hline
\!\!\!&\!\!\! ResNet-50 \!\!\!&\!\!\! ResNet-152 \!\!\!&\!\!\! VGG-16 \!\!\!& VAE-GAN\!\!\!\\
Raw filter~\cite{CNNSemanticDeep} & 0.1328 & 0.1346 & 0.1398 & 0.1944\\
Ours & {\bf0.0848} & {\bf0.0858} & {\bf0.0638} & {\bf0.1066}\\
\hline
{\footnotesize\cite{MiddleLevel}} & \multicolumn{4}{|c}{0.1341}\\
{\footnotesize\cite{CNNSemanticPart}} & \multicolumn{4}{|c}{0.2291}\\
\hline
\end{tabular}}
\caption{Location instability of patterns.}
\label{tab:stability}
\end{table}

\begin{figure}[t]
\centering
\includegraphics[width=0.9\linewidth]{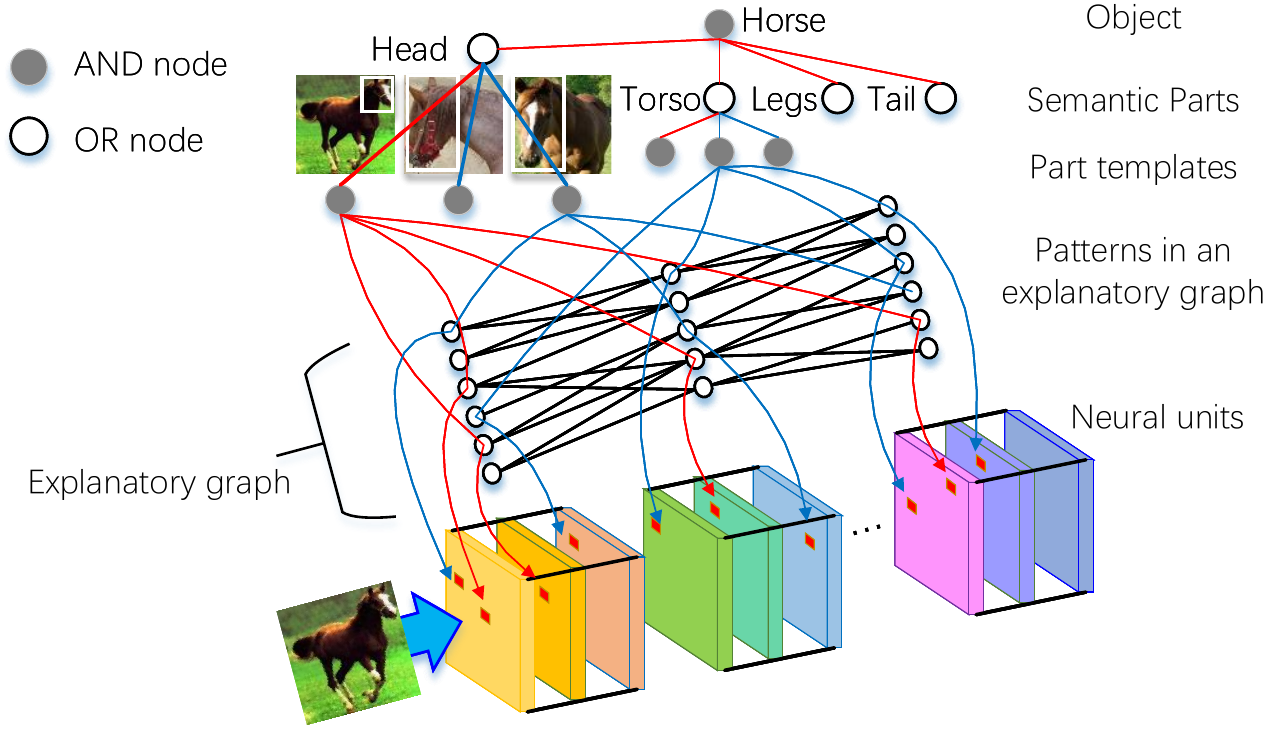}
\caption{And-Or graph for semantic object parts. The AOG encodes a four-layer hierarchy for each semantic part, \emph{i.e.} the semantic part (OR node), part templates (AND node), latent part patterns (OR nodes, those from the explanatory graph), and neural units (terminal nodes). In the AOG, the OR node of semantic part contains a number of alternative appearance candidates as children. Each OR node of a latent part pattern encodes a list of neural units as alternative deformation candidates. Each AND node (\emph{e.g.} a part template) uses a number of latent part patterns to describe its compositional regions.}
\label{fig:hybrid}
\end{figure}

\textbf{Location instability of inference positions:} We also defined the location instability of inference positions for each pattern as an alternative evaluation of pattern interpretability. We assumed that if a pattern was always triggered by the same object part through different images, then the distance between the pattern's inference position and a ground-truth landmark of the object part should not change a lot among various images.

As shown in Fig.~\ref{fig:instability}, for each testing image $I$, we computed the distances between the inferred position of $V$ and ground-truth landmark positions of \textit{head}, \textit{back}, and \textit{tail} parts, denoted by {\small$d_{I}^{\textrm{head}}$}, {\small$d_{I}^{\textrm{back}}$}, and {\small$d_{I}^{\textrm{tail}}$}. We normalized these distances by the diagonal length of input images. Then, we computed {\small$(\sqrt{var(d_{I}^{\textrm{head}})}+\sqrt{var(d_{I}^{\textrm{back}})}+\sqrt{var(d_{I}^{\textrm{tail}})})/3$} as the location instability of the node for evaluation, where {\small$var(d_{I}^{\textrm{head}})$} denotes the variation of {\small$d_{I}^{\textrm{head}}$} among different images.

Given an explanatory graph, we compared its location instability with three baselines. In the first baseline, we treated each filter in a CNN as a detector of a certain pattern. Thus, given the feature map of a filter (after the ReLu operation), we used the method of \cite{CNNSemanticDeep} to localize the unit with the highest response value as the pattern position. The other two baselines were typical methods to extract middle-level features from images~\cite{MiddleLevel} and extract patterns from CNNs~\cite{CNNSemanticPart}, respectively. For each baseline, we chose the top-500 patterns (\emph{i.e.} 500 nodes with top scores in our explanatory graph, 500 filters with strongest activations in the CNN, and the top-500 middle-level features). For each pattern, we selected position inferences on the top-20 images with highest scores to compute the instability of its inferred positions. Table~\ref{tab:stability} compares the location instability of the patterns learned by different baselines, and our method exhibited significantly lower location instability.

\subsection{Experiment 3: multi-shot part localization}

\begin{table}[t]
\centering
\resizebox{0.9\linewidth}{!}{\begin{tabular}{c|lcc}
\hline
&\multicolumn{2}{r}{Method$\qquad$obj.-box fine-tune}&\\
\multirow{3}{*}{\rotatebox[origin=c]{90}{no-RL}}
&{\small SS-DPM-Part~\cite{SSDPM}} & {N}
&{\small0.3469}
\\
&{\small PL-DPM-Part~\cite{PLDPM}} & {N}
&{\small0.3412}
\\
&{\small Part-Graph~\cite{SemanticPart}} & {N}
&{\small0.4889}
\\
\hline
\multirow{3}{*}{\rotatebox[origin=c]{90}{\scriptsize unsup\textcolor{red}{\footnotemark[5]}-RL}}
&{\small CNN-PDD~\cite{CNNSemanticPart}} & {N}
&{\small0.2333}
\\
&{\small CNN-PDD-ft~\cite{CNNSemanticPart}} & {Y}
&{\small0.3269}
\\
&{\small\bf Ours} & {Y}
&{\small\bf 0.0862}
\\
\hline
\multirow{4}{*}{\rotatebox[origin=c]{90}{sup-RL}}
&{\small fc7+linearSVM} & {Y}
&{\small0.3120}
\\
&{\small fc7+sp+linearSVM} & {Y}
&{\small0.3120}
\\
&{\small Fast-RCNN (1 ft)~\cite{FastRCNN}} & {N}
&{\small0.4517}
\\
&{\small Fast-RCNN (2 fts)~\cite{FastRCNN}} & {Y}
&{\small0.4131}
\\
\hline
\end{tabular}}
\caption{Normalized distance of part localization on the CUB200-2011 dataset~\cite{CUB200}. The second column indicates whether the baseline used all object-box annotations in the category to fine-tune a CNN.}
\label{tab:CUB}
\end{table}

\begin{table*}[t]
\centering
\resizebox{0.75\linewidth}{!}{\begin{tabular}{c|l|c|ccccccc}
\hline
&\multicolumn{2}{r|}{$\qquad\qquad$obj.-box fine-tune}& bird & cat & cow & dog & horse & sheep & \textcolor{blue}{\bf\large Avg.}\\
\multirow{3}{*}{\rotatebox[origin=c]{90}{no-RL}}
&{\small SS-DPM-Part~\cite{SSDPM}} \!&\! {N}
&{\small0.356}
&{\small0.270}
&{\small0.264}
&{\small0.242}
&{\small0.262}
&{\small0.286}
&\textcolor{blue}{\small0.280}
\\
&{\small PL-DPM-Part~\cite{PLDPM}} \!&\! {N}
&{\small0.294}
&{\small0.328}
&{\small0.282}
&{\small0.312}
&{\small0.321}
&{\small0.840}
&\textcolor{blue}{\small0.396}
\\
&{\small Part-Graph~\cite{SemanticPart}} \!&\! {N}
&{\small0.360}
&{\small0.208}
&{\small0.263}
&{\small0.205}
&{\small0.386}
&{\small0.500}
&\textcolor{blue}{\small0.320}
\\
\cline{1-3}
\multirow{3}{*}{\rotatebox[origin=c]{90}{\scriptsize unsup\textcolor{red}{\footnotemark[5]}-RL}}
&{\small CNN-PDD~\cite{CNNSemanticPart}} \!&\! {N}
&{\small0.301}
&{\small0.246}
&{\small0.220}
&{\small0.248}
&{\small0.292}
&{\small0.254}
&\textcolor{blue}{\small0.260}
\\
&{\small CNN-PDD-ft~\cite{CNNSemanticPart}} \!&\! {Y}
&{\small0.358}
&{\small0.268}
&{\small0.220}
&{\small0.200}
&{\small0.302}
&{\small0.269}
&\textcolor{blue}{\small0.269}
\\
&{\small\bf Ours} \!&\! {Y}
&{\small\bf 0.162}
&{\small\bf 0.130}
&{\small0.258}
&{\small\bf 0.137}
&{\small\bf 0.181}
&{\small\bf 0.192}
&\textcolor{blue}{\small\bf 0.177}
\\
\cline{1-3}
\multirow{4}{*}{\rotatebox[origin=c]{90}{sup-RL}}
&{\small fc7+linearSVM} \!&\! {Y}
&{\small0.247}
&{\small0.174}
&{\small0.251}
&{\small0.217}
&{\small0.261}
&{\small0.317}
&\textcolor{blue}{\small0.244}
\\
&{\small fc7+sp+linearSVM} \!&\! {Y}
&{\small0.247}
&{\small0.174}
&{\small\bf 0.249}
&{\small0.217}
&{\small0.261}
&{\small0.317}
&\textcolor{blue}{\small0.244}
\\
&{\small Fast-RCNN (1 ft)~\cite{FastRCNN}} \!&\! {N}
&{\small0.324}
&{\small0.324}
&{\small0.325}
&{\small0.272}
&{\small0.347}
&{\small0.314}
&\textcolor{blue}{\small0.318}
\\
&{\small Fast-RCNN (2 fts)~\cite{FastRCNN}} \!&\! {Y}
&{\small0.350}
&{\small0.295}
&{\small0.255}
&{\small0.293}
&{\small0.367}
&{\small0.260}
&\textcolor{blue}{\small0.303}
\\
\hline
\end{tabular}}
\caption{Normalized distance of part localization on the Pascal VOC Part dataset~\cite{SemanticPart}. The second column indicates whether the baseline used all object-box annotations in the category to fine-tune a CNN.}
\label{tab:VOC}
\centering
\resizebox{0.95\linewidth}{!}{\begin{tabular}{c|l|c|cccccccccccccccc}
\hline
&\multicolumn{2}{r|}{$\qquad\qquad$obj.-box fine-tune} \!\!&\!\! gold. \!\!&\!\! bird \!\!&\!\! frog \!\!&\!\! turt. \!\!&\!\! liza. \!\!&\!\! koala \!\!&\!\! lobs. \!\!&\!\! dog \!\!&\!\! fox \!\!&\!\! cat \!\!&\!\! lion \!\!&\!\! tiger \!\!&\!\! bear \!\!&\!\! rabb. \!\!&\!\! hams. \!\!&\!\! squi.\\
\multirow{3}{*}{\rotatebox[origin=c]{90}{no-RL}}&
\!\!\! {\small SS-DPM-Part} \!\!\!&\!\!\! {N}
\!\!\!&\!\!\!{\small0.297}
\!\!\!&\!\!\!{\small0.280}
\!\!\!&\!\!\!{\small0.257}
\!\!\!&\!\!\!{\small0.255}
\!\!\!&\!\!\!{\small0.317}
\!\!\!&\!\!\!{\small0.222}
\!\!\!&\!\!\!{\small0.207}
\!\!\!&\!\!\!{\small0.239}
\!\!\!&\!\!\!{\small0.305}
\!\!\!&\!\!\!{\small0.308}
\!\!\!&\!\!\!{\small0.238}
\!\!\!&\!\!\!{\small0.144}
\!\!\!&\!\!\!{\small0.260}
\!\!\!&\!\!\!{\small0.272}
\!\!\!&\!\!\!{\small0.178}
\!\!\!&\!\!\!{\small0.261}
\\
&\!\!\! {\small PL-DPM-Part} \!\!\!&\!\!\! {N}
\!\!\!&\!\!\!{\small0.273}
\!\!\!&\!\!\!{\small0.256}
\!\!\!&\!\!\!{\small0.271}
\!\!\!&\!\!\!{\small0.321}
\!\!\!&\!\!\!{\small0.327}
\!\!\!&\!\!\!{\small0.242}
\!\!\!&\!\!\!{\small0.194}
\!\!\!&\!\!\!{\small0.238}
\!\!\!&\!\!\!{\small0.619}
\!\!\!&\!\!\!{\small0.215}
\!\!\!&\!\!\!{\small0.239}
\!\!\!&\!\!\!{\small0.136}
\!\!\!&\!\!\!{\small0.323}
\!\!\!&\!\!\!{\small0.228}
\!\!\!&\!\!\!{\small0.186}
\!\!\!&\!\!\!{\small0.281}
\\
&\!\!\! {\small Part-Graph} \!\!\!&\!\!\! {N}
\!\!\!&\!\!\!{\small0.363}
\!\!\!&\!\!\!{\small0.316}
\!\!\!&\!\!\!{\small0.241}
\!\!\!&\!\!\!{\small0.322}
\!\!\!&\!\!\!{\small0.419}
\!\!\!&\!\!\!{\small0.205}
\!\!\!&\!\!\!{\small0.218}
\!\!\!&\!\!\!{\small0.218}
\!\!\!&\!\!\!{\small0.343}
\!\!\!&\!\!\!{\small0.242}
\!\!\!&\!\!\!{\small0.162}
\!\!\!&\!\!\!{\small0.127}
\!\!\!&\!\!\!{\small0.224}
\!\!\!&\!\!\!{\small0.188}
\!\!\!&\!\!\!{\small0.131}
\!\!\!&\!\!\!{\small0.208}
\\
\cline{1-3}
\multirow{3}{*}{\rotatebox[origin=c]{90}{\scriptsize unsup\textcolor{red}{\footnotemark[5]}-RL}}
&\!\!\! {\small CNN-PDD} \!\!\!&\!\!\! {N}
\!\!\!&\!\!\!{\small0.316}
\!\!\!&\!\!\!{\small0.289}
\!\!\!&\!\!\!{\small0.229}
\!\!\!&\!\!\!{\small0.260}
\!\!\!&\!\!\!{\small0.335}
\!\!\!&\!\!\!{\small0.163}
\!\!\!&\!\!\!{\small0.190}
\!\!\!&\!\!\!{\small0.220}
\!\!\!&\!\!\!{\small0.212}
\!\!\!&\!\!\!{\small0.196}
\!\!\!&\!\!\!{\small0.174}
\!\!\!&\!\!\!{\small0.160}
\!\!\!&\!\!\!{\small0.223}
\!\!\!&\!\!\!{\small0.266}
\!\!\!&\!\!\!{\small0.156}
\!\!\!&\!\!\!{\small0.291}
\\
&\!\!\! {\small CNN-PDD-ft} \!\!\!&\!\!\! {Y}
\!\!\!&\!\!\!{\small0.302}
\!\!\!&\!\!\!{\small0.236}
\!\!\!&\!\!\!{\small0.261}
\!\!\!&\!\!\!{\small0.231}
\!\!\!&\!\!\!{\small0.350}
\!\!\!&\!\!\!{\small0.168}
\!\!\!&\!\!\!{\small0.170}
\!\!\!&\!\!\!{\small0.177}
\!\!\!&\!\!\!{\small0.264}
\!\!\!&\!\!\!{\small0.270}
\!\!\!&\!\!\!{\small0.206}
\!\!\!&\!\!\!{\small0.256}
\!\!\!&\!\!\!{\small0.178}
\!\!\!&\!\!\!{\small0.167}
\!\!\!&\!\!\!{\small0.286}
\!\!\!&\!\!\!{\small0.237}
\\
&\!\!\! {\small\bf Ours} \!\!\!&\!\!\! {Y}
\!\!\!&\!\!\!{\small\bf 0.090}
\!\!\!&\!\!\!{\small\bf 0.091}
\!\!\!&\!\!\!{\small\bf 0.095}
\!\!\!&\!\!\!{\small 0.167}
\!\!\!&\!\!\!{\small\bf 0.124}
\!\!\!&\!\!\!{\small\bf 0.084}
\!\!\!&\!\!\!{\small\bf 0.155}
\!\!\!&\!\!\!{\small 0.147}
\!\!\!&\!\!\!{\small\bf 0.081}
\!\!\!&\!\!\!{\small\bf 0.129}
\!\!\!&\!\!\!{\small\bf 0.074}
\!\!\!&\!\!\!{\small\bf 0.102}
\!\!\!&\!\!\!{\small\bf 0.121}
\!\!\!&\!\!\!{\small\bf 0.087}
\!\!\!&\!\!\!{\small\bf 0.097}
\!\!\!&\!\!\!{\small\bf 0.095}
\\
\cline{1-3}
\multirow{4}{*}{\rotatebox[origin=c]{90}{sup-RL}}
&\!\!\! {\small fc7+linearSVM} \!\!\!&\!\!\! {Y}
\!\!\!&\!\!\!{\small0.150}
\!\!\!&\!\!\!{\small0.318}
\!\!\!&\!\!\!{\small0.186}
\!\!\!&\!\!\!{\small0.150}
\!\!\!&\!\!\!{\small0.257}
\!\!\!&\!\!\!{\small0.156}
\!\!\!&\!\!\!{\small0.196}
\!\!\!&\!\!\!{\small0.136}
\!\!\!&\!\!\!{\small0.101}
\!\!\!&\!\!\!{\small0.138}
\!\!\!&\!\!\!{\small0.132}
\!\!\!&\!\!\!{\small0.163}
\!\!\!&\!\!\!{\small0.122}
\!\!\!&\!\!\!{\small0.139}
\!\!\!&\!\!\!{\small0.110}
\!\!\!&\!\!\!{\small0.262}
\\
&\!\!\! {\small fc7+sp+linearSVM} \!\!\!&\!\!\! {Y}
\!\!\!&\!\!\!{\small0.150}
\!\!\!&\!\!\!{\small0.318}
\!\!\!&\!\!\!{\small0.186}
\!\!\!&\!\!\!{\small\bf 0.150}
\!\!\!&\!\!\!{\small0.254}
\!\!\!&\!\!\!{\small0.156}
\!\!\!&\!\!\!{\small0.196}
\!\!\!&\!\!\!{\small\bf 0.136}
\!\!\!&\!\!\!{\small0.101}
\!\!\!&\!\!\!{\small0.138}
\!\!\!&\!\!\!{\small0.132}
\!\!\!&\!\!\!{\small0.163}
\!\!\!&\!\!\!{\small0.122}
\!\!\!&\!\!\!{\small0.139}
\!\!\!&\!\!\!{\small0.110}
\!\!\!&\!\!\!{\small0.262}
\\
&\!\!\! {\small Fast-RCNN (1 ft)} \!\!\!&\!\!\! {N}
\!\!\!&\!\!\!{\small0.261}
\!\!\!&\!\!\!{\small0.365}
\!\!\!&\!\!\!{\small0.265}
\!\!\!&\!\!\!{\small0.310}
\!\!\!&\!\!\!{\small0.353}
\!\!\!&\!\!\!{\small0.365}
\!\!\!&\!\!\!{\small0.289}
\!\!\!&\!\!\!{\small0.363}
\!\!\!&\!\!\!{\small0.255}
\!\!\!&\!\!\!{\small0.319}
\!\!\!&\!\!\!{\small0.251}
\!\!\!&\!\!\!{\small0.260}
\!\!\!&\!\!\!{\small0.317}
\!\!\!&\!\!\!{\small0.255}
\!\!\!&\!\!\!{\small0.255}
\!\!\!&\!\!\!{\small0.169}
\\
&\!\!\! {\small Fast-RCNN (2 fts)} \!\!\!&\!\!\! {Y}
\!\!\!&\!\!\!{\small0.340}
\!\!\!&\!\!\!{\small0.351}
\!\!\!&\!\!\!{\small0.388}
\!\!\!&\!\!\!{\small0.327}
\!\!\!&\!\!\!{\small0.411}
\!\!\!&\!\!\!{\small0.119}
\!\!\!&\!\!\!{\small0.330}
\!\!\!&\!\!\!{\small0.368}
\!\!\!&\!\!\!{\small0.206}
\!\!\!&\!\!\!{\small0.170}
\!\!\!&\!\!\!{\small0.144}
\!\!\!&\!\!\!{\small0.160}
\!\!\!&\!\!\!{\small0.230}
\!\!\!&\!\!\!{\small0.230}
\!\!\!&\!\!\!{\small0.178}
\!\!\!&\!\!\!{\small0.205}
\\
\hline
&\!\!\!&\!\!\!&\!\!\! horse \!\!\!&\!\!\! zebra \!\!\!&\!\!\! swine \!\!\!&\!\!\! hippo \!\!\!&\!\!\! catt. \!\!\!&\!\!\! sheep \!\!\!&\!\!\! ante. \!\!\!&\!\!\! camel \!\!\!&\!\!\! otter \!\!\!&\!\!\! arma. \!\!\!&\!\!\! monk. \!\!\!&\!\!\! elep. \!\!\!&\!\!\! red pa. \!\!\!&\!\!\! gia.pa. \!\!\!&\!\!\! \!\!\!&\!\!\! \textcolor{blue}{\bf\large Avg.}\\
\multirow{3}{*}{\rotatebox[origin=c]{90}{no-RL}}
&\!\!\! {\small SS-DPM-Part} \!\!\!&\!\!\! {N}
\!\!\!&\!\!\!{\small0.246}
\!\!\!&\!\!\!{\small\bf 0.206}
\!\!\!&\!\!\!{\small0.240}
\!\!\!&\!\!\!{\small0.234}
\!\!\!&\!\!\!{\small0.246}
\!\!\!&\!\!\!{\small0.205}
\!\!\!&\!\!\!{\small0.224}
\!\!\!&\!\!\!{\small0.277}
\!\!\!&\!\!\!{\small0.253}
\!\!\!&\!\!\!{\small0.283}
\!\!\!&\!\!\!{\small0.206}
\!\!\!&\!\!\!{\small0.219}
\!\!\!&\!\!\!{\small0.256}
\!\!\!&\!\!\!{\small0.129}
\!\!\!&\!\!\!
\!\!\!&\!\!\!{\small\textcolor{blue}{0.242}}
\\
&\!\!\! {\small PL-DPM-Part} \!\!\!&\!\!\! {N}
\!\!\!&\!\!\!{\small0.322}
\!\!\!&\!\!\!{\small0.267}
\!\!\!&\!\!\!{\small0.297}
\!\!\!&\!\!\!{\small0.273}
\!\!\!&\!\!\!{\small0.271}
\!\!\!&\!\!\!{\small0.413}
\!\!\!&\!\!\!{\small0.337}
\!\!\!&\!\!\!{\small0.261}
\!\!\!&\!\!\!{\small0.286}
\!\!\!&\!\!\!{\small0.295}
\!\!\!&\!\!\!{\small0.187}
\!\!\!&\!\!\!{\small0.264}
\!\!\!&\!\!\!{\small0.204}
\!\!\!&\!\!\!{\small0.505}
\!\!\!&\!\!\!
\!\!\!&\!\!\!{\small\textcolor{blue}{0.284}}
\\
&\!\!\! {\small Part-Graph} \!\!\!&\!\!\! {N}
\!\!\!&\!\!\!{\small0.296}
\!\!\!&\!\!\!{\small0.315}
\!\!\!&\!\!\!{\small0.306}
\!\!\!&\!\!\!{\small0.378}
\!\!\!&\!\!\!{\small0.333}
\!\!\!&\!\!\!{\small0.230}
\!\!\!&\!\!\!{\small0.216}
\!\!\!&\!\!\!{\small0.317}
\!\!\!&\!\!\!{\small0.227}
\!\!\!&\!\!\!{\small0.341}
\!\!\!&\!\!\!{\small0.159}
\!\!\!&\!\!\!{\small0.294}
\!\!\!&\!\!\!{\small0.276}
\!\!\!&\!\!\!{\small0.094}
\!\!\!&\!\!\!
\!\!\!&\!\!\!{\small\textcolor{blue}{0.257}}
\\
\cline{1-3}
\multirow{3}{*}{\rotatebox[origin=c]{90}{\scriptsize unsup\textcolor{red}{\footnotemark[5]}-RL}}
&\!\!\! {\small CNN-PDD} \!\!\!&\!\!\! {N}
\!\!\!&\!\!\!{\small0.261}
\!\!\!&\!\!\!{\small0.266}
\!\!\!&\!\!\!{\small\bf 0.189}
\!\!\!&\!\!\!{\small0.192}
\!\!\!&\!\!\!{\small0.201}
\!\!\!&\!\!\!{\small0.244}
\!\!\!&\!\!\!{\small0.208}
\!\!\!&\!\!\!{\small0.193}
\!\!\!&\!\!\!{\small0.174}
\!\!\!&\!\!\!{\small0.299}
\!\!\!&\!\!\!{\small0.236}
\!\!\!&\!\!\!{\small0.214}
\!\!\!&\!\!\!{\small0.222}
\!\!\!&\!\!\!{\small0.179}
\!\!\!&\!\!\!
\!\!\!&\!\!\!{\small\textcolor{blue}{0.225}}
\\
&\!\!\! {\small CNN-PDD-ft} \!\!\!&\!\!\! {Y}
\!\!\!&\!\!\!{\small0.310}
\!\!\!&\!\!\!{\small0.321}
\!\!\!&\!\!\!{\small0.216}
\!\!\!&\!\!\!{\small0.257}
\!\!\!&\!\!\!{\small0.220}
\!\!\!&\!\!\!{\small0.179}
\!\!\!&\!\!\!{\small0.229}
\!\!\!&\!\!\!{\small0.253}
\!\!\!&\!\!\!{\small0.198}
\!\!\!&\!\!\!{\small0.308}
\!\!\!&\!\!\!{\small0.273}
\!\!\!&\!\!\!{\small0.189}
\!\!\!&\!\!\!{\small0.208}
\!\!\!&\!\!\!{\small0.275}
\!\!\!&\!\!\!
\!\!\!&\!\!\!{\small\textcolor{blue}{0.240}}
\\
&\!\!\! {\small\bf Ours} \!\!\!&\!\!\! {Y}
\!\!\!&\!\!\!{\small\bf 0.189}
\!\!\!&\!\!\!{\small 0.212}
\!\!\!&\!\!\!{\small 0.212}
\!\!\!&\!\!\!{\small 0.151}
\!\!\!&\!\!\!{\small\bf 0.185}
\!\!\!&\!\!\!{\small\bf 0.124}
\!\!\!&\!\!\!{\small\bf 0.093}
\!\!\!&\!\!\!{\small\bf 0.120}
\!\!\!&\!\!\!{\small\bf 0.102}
\!\!\!&\!\!\!{\small\bf 0.188}
\!\!\!&\!\!\!{\small\bf 0.086}
\!\!\!&\!\!\!{\small 0.174}
\!\!\!&\!\!\!{\small\bf 0.104}
\!\!\!&\!\!\!{\small\bf 0.073}
\!\!\!&\!\!\!
\!\!\!&\!\!\!{\small\textcolor{blue}{\bf 0.125}}
\\
\cline{1-3}
\multirow{4}{*}{\rotatebox[origin=c]{90}{sup-RL}}
&\!\!\! {\small fc7+linearSVM} \!\!\!&\!\!\! {Y}
\!\!\!&\!\!\!{\small0.205}
\!\!\!&\!\!\!{\small0.258}
\!\!\!&\!\!\!{\small0.201}
\!\!\!&\!\!\!{\small0.140}
\!\!\!&\!\!\!{\small0.256}
\!\!\!&\!\!\!{\small0.236}
\!\!\!&\!\!\!{\small0.164}
\!\!\!&\!\!\!{\small0.190}
\!\!\!&\!\!\!{\small0.140}
\!\!\!&\!\!\!{\small0.252}
\!\!\!&\!\!\!{\small0.256}
\!\!\!&\!\!\!{\small0.176}
\!\!\!&\!\!\!{\small0.215}
\!\!\!&\!\!\!{\small0.116}
\!\!\!&\!\!\!
\!\!\!&\!\!\!{\small\textcolor{blue}{0.184}}
\\
&\!\!\! {\small fc7+sp+linearSVM} \!\!\!&\!\!\! {Y}
\!\!\!&\!\!\!{\small0.205}
\!\!\!&\!\!\!{\small0.258}
\!\!\!&\!\!\!{\small0.201}
\!\!\!&\!\!\!{\small\bf 0.140}
\!\!\!&\!\!\!{\small0.256}
\!\!\!&\!\!\!{\small0.236}
\!\!\!&\!\!\!{\small0.164}
\!\!\!&\!\!\!{\small0.190}
\!\!\!&\!\!\!{\small0.140}
\!\!\!&\!\!\!{\small0.250}
\!\!\!&\!\!\!{\small0.256}
\!\!\!&\!\!\!{\small0.176}
\!\!\!&\!\!\!{\small0.215}
\!\!\!&\!\!\!{\small0.116}
\!\!\!&\!\!\!
\!\!\!&\!\!\!{\small\textcolor{blue}{0.184}}
\\
&\!\!\! {\small Fast-RCNN (1 ft)} \!\!\!&\!\!\! {N}
\!\!\!&\!\!\!{\small0.374}
\!\!\!&\!\!\!{\small0.322}
\!\!\!&\!\!\!{\small0.285}
\!\!\!&\!\!\!{\small0.265}
\!\!\!&\!\!\!{\small0.320}
\!\!\!&\!\!\!{\small0.277}
\!\!\!&\!\!\!{\small0.255}
\!\!\!&\!\!\!{\small0.351}
\!\!\!&\!\!\!{\small0.340}
\!\!\!&\!\!\!{\small0.324}
\!\!\!&\!\!\!{\small0.334}
\!\!\!&\!\!\!{\small0.256}
\!\!\!&\!\!\!{\small0.336}
\!\!\!&\!\!\!{\small0.274}
\!\!\!&\!\!\!
\!\!\!&\!\!\!{\small\textcolor{blue}{0.299}}
\\
&\!\!\! {\small Fast-RCNN (2 fts)} \!\!\!&\!\!\! {Y}
\!\!\!&\!\!\!{\small0.346}
\!\!\!&\!\!\!{\small0.303}
\!\!\!&\!\!\!{\small0.212}
\!\!\!&\!\!\!{\small0.223}
\!\!\!&\!\!\!{\small0.228}
\!\!\!&\!\!\!{\small0.195}
\!\!\!&\!\!\!{\small0.175}
\!\!\!&\!\!\!{\small0.247}
\!\!\!&\!\!\!{\small0.280}
\!\!\!&\!\!\!{\small0.319}
\!\!\!&\!\!\!{\small0.193}
\!\!\!&\!\!\!{\small\bf 0.125}
\!\!\!&\!\!\!{\small0.213}
\!\!\!&\!\!\!{\small0.160}
\!\!\!&\!\!\!
\!\!\!&\!\!\!{\small\textcolor{blue}{0.246}}
\\
\hline
\end{tabular}}
\caption{Normalized distance of part localization on the ILSVRC 2013 DET Animal-Part dataset~\cite{CNNAoG}. The second column indicates whether the baseline used all object-box annotations in the category to fine-tune a CNN.}
\label{tab:imgnet}
\end{table*}

\subsubsection{And-Or graph for semantic parts}

The explanatory graph makes it plausible to transfer middle-layer patterns from CNNs to semantic object parts. In order to test the transferability of patterns, we build an additional And-Or graph (AOG) to associate certain implicit patterns with an explicit part name, in the scenario of multi-shot learning. We used the AOG to localize semantic parts of objects for evaluation. The structure of the AOG is inspired by \cite{NineAOG}, and the learning of the AOG was originally proposed in \cite{CNNAoG}. We briefly introduce the AOG in \cite{CNNAoG} as follows.

As shown in Fig.~\ref{fig:hybrid}, like the hierarchical model in \cite{HierarchicalFace}, the AOG encodes a four-layer hierarchy for each semantic part, \emph{i.e.} the semantic part (OR node), part templates (AND node), latent patterns (OR nodes, those from the explanatory graph), and neural units (terminal nodes). In the AOG, each OR node (\emph{e.g.} a semantic part or a latent pattern) contains a list of alternative appearance (or deformation) candidates. Each AND node (\emph{e.g.} a part template) uses a number of latent patterns to describe its compositional regions.

1) The OR node of a semantic part contains a total of $m$ part templates to represent alternative appearance or pose candidates of the part. 2) Each part template (AND node) retrieve $K$ patterns from the explanatory graph as children. These patterns describe compositional regions of the part. 3) Each latent pattern (OR node) has all units in its corresponding filter's feature map as children, which represent its deformation candidates on image $I$.

\subsubsection{Experimental settings of three-shot learning}

We learned the explanatory graph based on a fine-tuned VGG-16 network and built the AOG following the scenario of multi-shot learning introduced in \cite{CNNAoG}. For each category, we used three annotations of the head part to learn three head templates in the AOG. Such part annotations were offered by \cite{CNNAoG}. To enable a fair comparison, all the object-box annotations and the three part annotations were equally provided to all baselines for learning.

We learned the explanatory graph based on a fine-tuned VGG-16 network~\cite{VGG} and built the AOG following the scenario of multi-shot learning introduced in \cite{CNNAoG}. For each category, we set three templates for the head part ($m=3$), and used a single part-box annotation for each template. We set {\small$K\!=\!0.1\sum_{L,d}N_{L,d}$} to learn AOGs for categories in the ILSVRC Animal-Part and CUB200 datasets and set {\small$K\!=\!0.4\sum_{L,d}N_{L,d}$} for Pascal VOC Part categories. Then, we used the AOGs to localize semantic parts on objects. Note that we used object images without part annotations to learn the explanatory graph and we used three part annotations provided by \cite{CNNAoG} to build the AOG. All these training samples were equally provided to all baselines for learning (besides part annotations, all baselines also used object annotations contained in the datasets for learning).

\textbf{Baselines:}{\verb| |} We compared AOGs with a total of ten baselines in part localization. The baselines included 1) state-of-the-art algorithms for object detection (\emph{i.e.} directly detecting target parts from objects), 2) graphical/part models for part localization, and 3) the methods selecting CNN patterns to describe object parts.

The first baseline was the standard fast-RCNN~\cite{FastRCNN}, namely \textit{Fast-RCNN (1 ft)}, which directly fine-tuned a VGG-16 network based on part annotations. Then, the second baseline, namely \textit{Fast-RCNN (2 fts)}, first used massive object-box annotations in the target category to fine-tune the VGG-16 network with the loss of object detection. Then, given part annotations, Fast-RCNN (2 fts) further fine-tuned the VGG-16 to detect object parts. We used \cite{CNNSemanticPart} as the third baseline, namely \textit{CNN-PDD}. CNN-PDD selected certain filters of a CNN to localize the target part. In CNN-PDD, the CNN was pre-trained using the ImageNet dataset~\cite{ImageNet}. Just like Fast-RCNN (2 ft), we extended \cite{CNNSemanticPart} as the fourth baseline \textit{CNN-PDD-ft}, which fine-tuned a VGG-16 network using object-box annotations before applying the technique of \cite{CNNSemanticPart}. The fifth and sixth baselines were DPM-related methods, \emph{i.e.} the strongly supervised DPM (\textit{SS-DPM-Part})~\cite{SSDPM} and the technique in \cite{PLDPM} (\textit{PL-DPM-Part}), respectively. Then, the seventh baseline, namely \textit{Part-Graph}, used a graphical model for part localization~\cite{SemanticPart}. For weakly supervised learning, ``simple'' methods are usually insensitive to model over-fitting. Thus, we designed two baselines as follows. First, we used object-box annotations in a category to fine-tune the VGG-16 network. Then, given a few well-cropped object images, we used the selective search~\cite{SelectiveSearch} to collect image patches, and used the VGG-16 network to extract \textit{fc7} features from these patches. The baseline \textit{fc7+linearSVM} used a linear SVM to detect the target part. The other baseline \textit{fc7+sp+linearSVM} combined both the \textit{fc7} feature and the spatial position {\small$(x,y)$ ($-1\leq x,y\leq1$)} of each image patch as features for part detection. The last competing method is weakly supervised mining of part patterns from CNNs~\cite{CNNAoG}, namely \textit{supervised-AOG}. Unlike our method (unsupervised), \textit{supervised-AOG} used part annotations to extract part patterns.

\begin{table}[t]
\centering
\resizebox{1.0\linewidth}{!}{\begin{tabular}{l|ccc}
\hline
{\small Dataset} & {\scriptsize ILSVRC DET Animal} & {\scriptsize Pascal VOC Part} & {\scriptsize CUB200-2011}\\
\hline
{\small Supervised-AOG} & 0.1344 & 0.1767 & 0.0915\\
{\small Ours (unsupervised)} & {\bf0.1250} & {\bf0.1765} & {\bf0.0862}\\
\hline
\end{tabular}}
\caption{Normalized distance of part localization. We compared supervised and unsupervised mining of part patterns.}
\label{tab:cnnaog}
\end{table}

\textbf{Comparisons:}{\verb| |} To enable a fair comparison, we classify all baselines into three groups, \emph{i.e.} no representation learning (no-RL), unsupervised representation learning (unsup-RL)\footnote[5]{Representation learning in these methods only used object-box annotations, which is independent to part annotations. A few part annotations were used to select off-the-shelf pre-trained features.}, and supervised representation learning (sup-RL). The No-RL group includes conventional methods without using deep features, such as SS-DPM-Part, PL-DPM-Part, and Part-Graph. Sup-RL methods are Fast-RCNN (1 ft), Fast-RCNN (2 ft), CNN-PDD, CNN-PDD-ft, supervised-AOG, fc7+linearSVM, and fc7+sp+linearSVM. Fast-RCNN methods used part annotations to learn features. Supervised-AOG used part annotations to select filters from CNNs to localize parts. Unsup-RL methods include CNN-PDD, CNN-PDD-ft, and our method. These methods did not use part annotations, and only used object boxes for learning/selection.

We use the normalized distance to evaluate localization accuracy, which has been used in \cite{CNNAoG,CNNSemanticPart} as a standard metric. Tables~\ref{tab:CUB}, \ref{tab:VOC}, and \ref{tab:imgnet} show part-localization results on the CUB200-2011 dataset~\cite{CUB200}, the Pascal VOC Part dataset~\cite{SemanticPart}, and the ILSVRC 2013 DET Animal-Part dataset~\cite{CNNAoG}, respectively. Table~\ref{tab:cnnaog} compares the unsupervised and supervised learning of neural patterns. In the experiment, the AOG outperformed all baselines, even methods that learned part features in a supervised manner.

\section{Conclusion and discussions}

In this paper, we proposed a simple yet effective method to learn an explanatory graph that reveals knowledge hierarchy inside conv-layers of a pre-trained CNN (\emph{e.g.} a VGG-16, a residual network, or a VAE-GAN). We regard the graph as a concise and meaningful representation, which 1) filters out noisy activations, 2) disentangles reliable part patterns from each filter of the CNN, and 3) encodes co-activation logics and spatial relationships between patterns. Experiments showed that our patterns had significantly higher stability than baselines.

The explanatory graph's transparent representation makes it plausible to transfer CNN patterns to object parts. Part-localization experiments well demonstrated the good transferability. Our method even outperformed supervised learning of part representations. Nevertheless, the explanatory graph is still a rough representation of the CNN, rather than an accurate reconstruction of the CNN knowledge.

\section*{Acknowledgement}
This work is supported by ONR MURI project N00014-16-1-2007 and DARPA XAI Award N66001-17-2-4029, and NSF IIS 1423305.

{\small
\bibliographystyle{aaai}
\bibliography{TheBib}
}

\end{document}